\documentclass[journal]{IEEEtran}
\usepackage{threeparttable}
\usepackage{float}
\restylefloat{table}
\usepackage{stfloats}
\usepackage{diagbox}
\usepackage{multirow}
\usepackage{soul}
\usepackage{url}

\usepackage{graphicx}
\usepackage{amsmath}
\usepackage{booktabs}
\usepackage{indentfirst}
\ifCLASSINFOpdf
\else
\fi
\begin{document}
%
\title{Aesthetic Photo Collage with Deep Reinforcement Learning}
%
%
%

\author{Mingrui Zhang,
        Mading Li,
        Li Chen, 
        Jiahao Yu}
\maketitle

\begin{abstract}
    Photo collage aims to automatically arrange multiple photos on a given canvas with high aesthetic quality. Existing methods are based mainly on handcrafted feature optimization, which cannot adequately capture high-level human aesthetic senses. Deep learning provides a promising way, but owing to the complexity of collage and lack of training data, a solution has yet to be found. In this paper, we propose a novel pipeline for automatic generation of aspect ratio specified collage and the reinforcement learning technique is introduced in collage for the first time. Inspired by manual collages, we model the collage generation as a sequential decision process to adjust spatial positions, orientation angles, placement order and the global layout. To instruct the agent to improve both the overall layout and local details, the reward function is specially designed for collage, considering subjective and objective factors. To overcome the lack of training data, we pretrain our deep aesthetic network on a large scale image aesthetic dataset (CPC) for general aesthetic feature extraction and propose an attention fusion module for structural collage feature representation. We test our model against competing methods on two movie datasets and our results outperform others in aesthetic quality evaluation. A further user study is also conducted to demonstrate the effectiveness. 
    
\end{abstract}


%
\IEEEpeerreviewmaketitle

\section{Introduction}
With the rapid development of the Internet, there has been an increasing popularity of multimedia. The steep surge of images and videos has lead to information explosion and how to efficiently display diverse content to users within limited spaces has become a popular topic.

Photo collage has been proposed to automatically arrange multiple images on a given canvas. It is widely used for various purposes, such as advertising and personal photo summarization.  Yet the scalability and flexibility also make it a challenging task to generate photo collage with high aesthetic quality.

In past decades, many works have been published addressing this issue. One method for solving this problem is based on canvas partitioning, which typically involves circle packing algorithm \cite{yu2013content} or feature embedding \cite{liu2017correlation}  to partition the canvas into separate disjointed areas. Another method employs customized features \cite{rother2006autocollage,liu2009picture} to assess the quality of a collage and minimize the energy term through complex optimizations.

However, such handcrafted-based methods cannot provide adequate collage representation and generate high-quality collages of general scenes. To overcome the limitations of handcrafted features, deep learning can provide comprehensive feature representation. However, owing to the subjectivity and complexity of collage tasks, training data is lacking and unsuitable for supervised learning. Moreover, photo collage is a multistep task, which increases the infeasibility of directly applying deep learning.

Motivated by these challenges, we propose a novel pipeline for automatic photo collage generation. Inspired by manual collages, we decompose the collage generation into interpretable steps and model it as a reinforcement learning (RL) process for the first time. 
As illustrated in Figure \ref{fig:network}, the proposed model includes the deep aesthetic network and the collage generation module. 

  Since manual annotations for photo collage require highly skilled designers, and photo collages are complex and need a substantial amount of training data; thus, the high cost involved makes constructing a training dataset for photo collages unrealistic. 
  Consequently, we pretrain our aesthetic network on a large scale image aesthetic dataset (i.e., Comparative Photo Composition (CPC)) \cite{wei2018good} for general aesthetic feature extraction and propose an attention fusion module for collage feature extraction. 
  The attention fusion module is designed to extract the complex structural features of a photo collage, such as the composition and collocation of different subimages. 
   Specifically, for better adaptation in photo collage, the attention fusion module adopts multi-patch information with an attention mechanism to effectively represent the complex features of a collage, inspired by image aesthetic evaluation\cite{lu2014rapid,lu2015deep,sheng2018attention}.
    

   With the aesthetic and structural feature representation from deep aesthetic network, we formulate collage generation as a sequential decision process and present an improved RL framework. Speciﬁcally, we design a policy network to manipulate the global layout and local detail properties of individual images, including the orientation angle, relative spatial position, and placement order. In each step, the policy network makes improvements and generates an aspect ratio-speciﬁed collage, while the value network assists in stable and precise policy making. The reward design considers subjective and objective factors to instruct the agent to generate collage results with a balanced composition and less blank spaces. The policy making training is adapted to the advantage actor–critic (A2C) algorithm.
 
 In summary, our main contributions are as follows: 
\begin{itemize}
\item  We propose a novel pipeline for automatic photo collage generation. We decompose the collage generation into interpretable steps and model it as an reinforcement learning process, which to our knowledge, is the first work for directly applying deep learning in automatic photo collage.
\item  We develop the deep aesthetic network for general aesthetic feature extraction and propose an attention fusion module for structural collage feature representation. To overcome the lack of collage dataset, we pretrain the main aesthetic network on large scale image aesthetic dataset (CPC).
\item  We evaluate our model against several competing methods on the Hollywood2 and LSMDC3 movie dataset. Our model outperforms the other methods in aesthetic quality evaluation. To demonstrate the effectiveness of our model and the subjective visual quality, we conduct a user study.
\end{itemize}


\section{Related Works}
    \textbf{Photo collage}  aims to create a visually appealing summary by arranging multiple images on a given canvas. Previous works on photo collages mainly fall into three categories. 
    (a.) Region partitioning-based methods \cite{yu2013content,liu2017correlation} involve a circle packing algorithm \cite{lu2014rapid} or feature embedding \cite{liu2017correlation} to partition the canvas into separate disjointed areas.
    (b.) Content preserving collages \cite{atkins2008blocked,wu2013picwall} rely on tree-based page division to recursively split a canvas but ignore the image content. 
   (c.)  Customized energy terms optimizations-based methods \cite{rother2006autocollage,liu2009picture,goferman2010puzzle} cast photo collage as an optimization problem with well-defined objective functions.
    Compared with traditional methods which focus mainly on optimizing regions of interest and salience information, our method exploits deep learning technique to provide comprehensive photo collage representation, which in turn can beneﬁt the output of high-quality collages.
    
  \textbf{Aesthetic evaluation} for image is extensively examined and successfully employed in multiple tasks like image quality assessment \cite{kao2017deep,zeng2019unified}, image cropping \cite{wang2017deep,zeng2019reliable} and image composition \cite{chen2017learning,wei2018good}, benefiting from the powerful feature representation of deep neural networks.
  Typical image aesthetic assessment approaches rely on multi-patch representation \cite{lu2014rapid,lu2015deep,ma2017lamp,sheng2018attention}, which represents each image with multiple cropped patches to learn global and local detail information simultaneously and are proven to be useful.
   Although many succeeding works \cite{ma2017lamp,sheng2018attention} focus on improvement and further generalization, they remain limited to single images. By contrast, our deep aesthetic network is designed to extract general image aesthetic features and meaningful structural features for a photo collage with the help of the proposed attention fusion module, which can provide comprehensive photo collage representation.

\textbf{Reinforcement learning} methods were applied to multiple computer vision tasks in recent years, including image cropping \cite{li2018a2,li2019fast}, image enhancement \cite{park2018distort} image restoration \cite{yu2018crafting} and object tracking \cite{zhong2018hierarchical}. Such works simulate iterative manual modiﬁcation heuristically, making the operating steps interpretable and easy to understand. Compared with supervised methods, RL-based models do not require heavy annotations and are suitable for subjective tasks, such as collage generation.

\section{Methods}
\begin{figure*}
  \includegraphics[width=\textwidth,height=4.8cm]{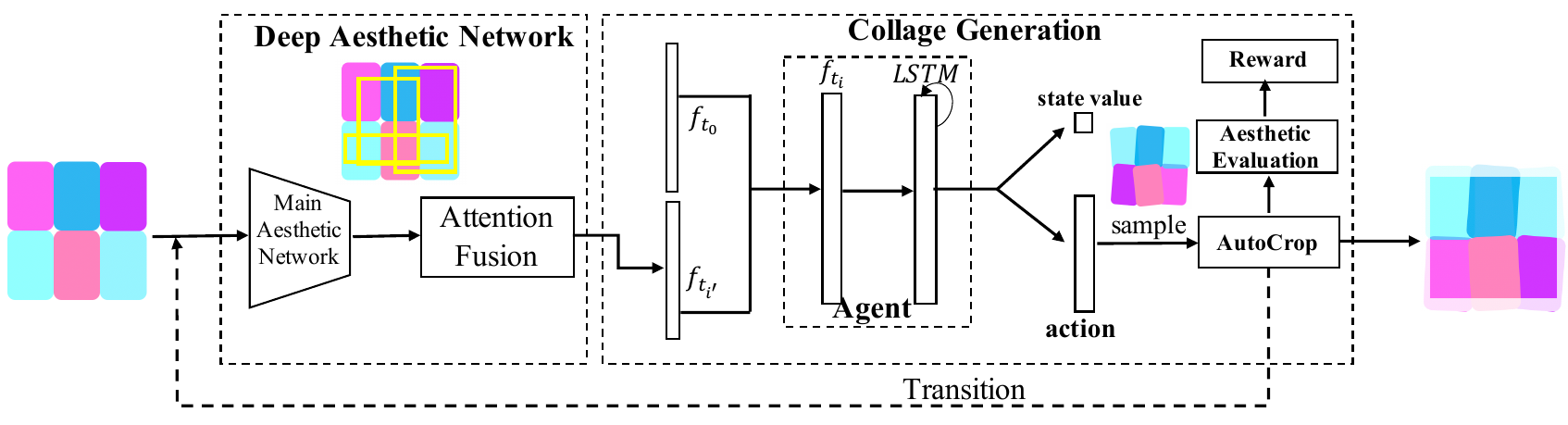}
  \vspace{1mm}
  \caption{The proposed network architecture for automatic collage generation.  First, the aesthetic network takes in concatenated pictures best suited for the aspect ratio specified canvas as initialization. At each step, the deep aesthetic network first extracts both general aesthetic feature and structural feature representations using the pretrained image aesthetic network and Attention Fusion module for the current collage. 
  Then the current feature concatenated with initial feature is fed into the actor-critic network with action and state output. The agent makes policy based on past observations and samples from the action space to manipulate the global layout of a collage and the local detail properties of individual images as described in Table \ref{tab:action}. Lastly, the AutoCrop module adapts the generated photo collage with irregular shapes to an aspect ratio-specified canvas, after which the evaluation network computes the aesthetic score and the reward is estimated for current policy.
  }\label{fig:network}
\end{figure*}

  Inspired by manual collages, the collage generation is decomposed into interpretable steps and modeled as reinforcement learning process for the first time. Figure \ref{fig:network} illustrates that the deep RL model includes a deep aesthetic network for comprehensive feature representation and a collage generation module for agent training.

\subsection{Deep Aesthetic Network for Photo Collage}
  
    The deep aesthetic network is designed to extract representative features for a collage.  It is composed of the main aesthetic network and the attention fusion module for comprehensive collage feature representation.
    
    \textbf{Main aesthetic network for general aesthetic feature extraction}. Since a collage is composed of multiple images, encoding information directly from the holistic collage may cause vast information loss and cannot capture local details. Alternatively, the main aesthetic network represents the collage with a bag of  predefined patches that densely slides over different scales and aspect ratios of a normalized collage, aiming to explore general aesthetic attributes among images.
    
    The network architecture is composed of nine layer convolution blocks, resembling the object detection architecture Single Shot MultiBox Detector\cite{liu2016ssd}. 
    Due to high expense of manual annotations and lack of high quality photo collage datasets, we pretrain our main aesthetic network on a large scale image aesthetic dataset (CPC dataset) for general aesthetic feature representation.

     
     \textbf{Attention Fusion Module for structural aesthetic feature representation.} The components include the fusion module and the attention layer. The fusion module provides aggregated information from orderless patches for better adaptation in collage. On this basis, the attention layer is designed for structural aesthetic feature representation of a collage.
     
    Since the most concerned part for a collage is the composition and color collocation of composed subimages, the fusion module shifts more focus to the composition among adjacent subimages instead of local parts from single image.  Specifically, the fusion module set standards for feature selection extracted from a candidate patch with an area proportion greater than $\eta $  in order to discover better composition quality and more harmonious content placement, which is calculated as follows:
\begin{equation}
\small
 f_{p_{i}}^{\prime}=f_{p_{i}} \cdot \delta\left(\frac{s\left(P_{i}\right)}{s(C)}>\eta\right) 
\end{equation}
where $P_i$  stands for $i$-th patch, $s(\cdot)$  and $f$ stands for the area and feature for the $i$-th patch and $\delta(x)=1$ if $x$ is true. And the collage is represented as $f(C)=\left[f_{p_{1}}^{\prime}, f_{p_{2}}^{\prime}, \ldots, f_{p_{n}}^{\prime}\right]$.


  The attention layer assigns dynamic weights to features from selected patches for effective learning of the complex structural features of a collage.
  Similar to handcrafted collages which highlight the important images, we introduce the Rule of Central frequently used in photography and pay specific attention to patches close to the center of the collage for further aesthetic quality improvement.  
  Since the patch information is extracted sharing the same weights, we inherently add the central rules in the multi-patch fusion process via attention mechanism, which is formulated as:

   \begin{equation}
   \small
    l_{i} = \|\frac{y_{i}}{h_{c}}-y_{c}\|_2 + \|\frac{x_{i}}{w_{c}}-x_{c}\|_2
   \end{equation}
\begin{equation}
\small
\alpha_{i}= \frac{s(P_{i})}{s(C)} \cdot(1- l_{i})
\end{equation}
$(y_{i}, x_{i})$  denotes the center coordinate of $i$-th patch, $(h_{c}, w_{c})$ correspond to the canvas size, and  $(y_{c}, x_{c})$  is chosen (0.5, 0,5) to represent the focus center on canvas. The weighting factors is therefore represented as $\alpha(C)=[\alpha_{1}, \alpha_{2}, \ldots, \alpha_{n}]$. Finally, combining the general and structural aesthetic feature, the representation of the holistic collage is computed as:
\begin{equation}
\small
F(C)=\alpha(C) \cdot f(C)
\end{equation}

\subsection{Collage Generation}

  We cast the collage generation as a sequential decision process and introduce the RL framework,where the overall learning target is to find the best global layout and the most adequate local details.
  
    Figure \ref{fig:network} depicts the pipeline and how the overall learning for collage generation is designed from state space, action space and reward function. The state and action space consider the global layout and local details, whereas the reward function is designed considering subjective and objective factors. The overall optimization goal is to maximize the accumulated reward of the trajectory generated by the agent’s policy.

\subsubsection{State and Action Space}\

  The agent keeps interacting with the environment by observing from the current state $s \in S$ of the environment and performs actions $a\in A$ according to the policy $\pi(a|s)$ .
    
    The observation $o_{t}$  includes the current aesthetic feature extracted from the collage concatenated with the initial feature. The state $s_{t} = \{o_{0},o_{1},...,o_{t}\}$  includes all the past observations. To take advantage of the historical experience, a long short-term memory (LSTM) unit is added in the agent, assisting in making a better policy. 
    
     Inspired from the quick initialization in \cite{liu2009picture} which optimizes the collage from layout to details, we customize the action space for collage generation into two categories to adjust the global layout and local details with different attributes.

  The global layout has substantial influences on the composition quality and is essential for a visually satisfying collage. 
    Inspired from long distance image dragging in manual collages, our agent switches one image pair at each step before reaching the max step. The switch action could be interpreted as changing the order of input images. It affects the global layout through optimizations of image adjacency collocations and is done in multiple consecutive steps to adjust positions of images of more importance(like group photos) in a collage.
    Additionally, as the agent is expected to decide the best layout when the score no longer increases, the termination action is designed as a trigger to stop the transforming process and output the current layout.
    
    For detail adjustments, the agent operates on each individual image. Inspired from the state variables defined on the image set in \cite{liu2009picture}, the detail adjustment attributes are designed with spatial position, placement order and orientation angles.
    
    Due to the complex search space of spatial search space, it is clearly inefficient to directly operate the coordinates of each individual image. 
    Alternatively, we formulate the learning of absolute spatial positions of separate images as relative displacement of adjacent images.  The overlay area between images is initialized as zero and altered progressively within one episode. 
    Subsequently, the agent adjusts the position of each image by reference to the neighboring image.
    The operations for layer are useful to display highlights and hide irrelevant area, while the rotation operation is designed to satisfy natural preference and improve the visual impression of the collage results.


\begin{table}[]

\begin{threeparttable}
\begin{tabular}{|c|c|c|}
\hline
\multicolumn{1}{|c|}{\textbf{Category}}                                & \textbf{Attribute}  & \textbf{Operation}  \\ \hline
\multirow{2}{*}{\begin{tabular}[c]{@{}c@{}}global layout \\ (C1) \end{tabular}} & layout              & switch image-pair \\ \cline{2-3} 
                                                                         & termination         & -                 \\ \hline
\multirow{4}{*}{\begin{tabular}[c]{@{}c@{}}local details \\ (C2) \end{tabular}} & x-relative-position & -15/-5/0/+5   (pixel)     \\ \cline{2-3} 
                                                                         & y-relative-position & -15/-5/0/+5 (pixel)       \\ \cline{2-3} 
                                                                         & layer               & top \tnote{1} /bottom \tnote{1} / -      \\ \cline{2-3} 
                                                                         & orientation angle   & -0.5/0/+0.5 (°)     \\ \hline
\end{tabular}

\begin{tablenotes}
        \footnotesize
        \item[1] The "top" (or "bottom") operation puts an image on the top (or bottom) layer to highlight (or hide) it.
\end{tablenotes}
\end{threeparttable}
\caption{The action space design considers global layout and local details. The global layout actions aim to improve the composition quality through image pair switching before the termination or max step. The local detail actions include relative spatial position, orientation angle and placement order (layer) for fine-grained collocation of sub-images. }
\label{tab:action}
\end{table}


\subsubsection{AutoCrop Module}\

  The AutoCrop Module adapts a collage with irregular shapes to the aspect ratio speciﬁed canvas during each episode. It incorporates the aspect ratio information into the environment and gives feedback to the agent at each step.
  To be specific, after the agent adjusts the collage at each step, multiple candidate views are cropped from the current collage, resulting in consistency with the canvas. Then the view selection is completed in the evaluation network and the cropped collage with the highest score is transitioned to the next step. 
   As a result, the agent is encouraged to choose actions that progressively avoids losing salient information and meanwhile suppress the blank space on the canvas.

\subsubsection{Reward Design}\
 
  The reward function describes preference for the current state and the overall target is to find the most visually pleasing collage result. To achieve this, the reward function considers subjective and objective scores to provide the agent incentive to optimize the collage quality at each step.
    
    Since the major challenge for analysing photo collage lies in the structural complexity, we assess the subjective quality with composition quality. Specifically, we resort to aesthetic evaluation network and propose using the aesthetic proposal number satisfying the aesthetic selection standard as representation for the subjective score of the collage $C$, which is denoted as $s_{a}(C)$. Intuitively, collage results with more aesthetic box number are indicators of higher aesthetic quality. For the objective score, since the salience is implicitly optimized by the aesthetic network, we calculate the blank area $s_{b}$ on the canvas. Overall, the evaluation score for a collage is computed as:

 

\begin{equation}
\small
s(C_{t})=\lambda_{a} s_{a}(C_{t})-\lambda_{b} s_{b}(C_{t})
\end{equation}
where the $\lambda_{a}$ and $\lambda_{b}$ stands for the weights for the aesthetic and the blank loss item, respectively. 
  After each step,  the difference of the aesthetic score between the updated collage  $C_{t+1}$ and the previous collage $C_{t}$  is used to calculate the reward for the current policy.  In order to increase the aesthetic quality whilst suppress the blank space, the agent is granted a positive reward if the score increases and a negative reward otherwise.
\begin{equation}
\small
r_{t}^{'}\left(C_{t}\right)=s(C_{t+1})-s(C_{t})
\end{equation}
  
    Finally, the agent is facilitated with the greedy strategy to avoid redundant actions and speed up generation progress because the reinforcement reward scheme indirectly treats the number of steps as a potential cost.
\begin{equation}
\small
r_{t}(C_{t}) =r_{t}^{'}(C_{t}) -0.01 *(t+1)
\end{equation}

\subsubsection{Training Algorithm}\

  We adopt the A2C algorithm as our RL framework to train the policy of collage generation. The A2C includes two sub-networks. The policy network $\theta_{p}$  outputs the probability distribution over the designed action space, each corresponding with the action operating on the collage according to the policy $\pi(a^{(t)} |s^{(t)})$. The value network outputs $V(s_{t};\theta_{v})$, which predicts the expected accumulated reward $R_{t}$ at step $t$.. Both networks share the backbone to reduce parameters. The global reward $R_{t}$ is estimated as $ r^{t} + \gamma V(s^{t})$ and the overall optimization target during training is described as follows:     
    \begin{equation}
    \small
    L_{\theta_{p}}=-\log \pi(a \mid s^{(t)})(R^{(t)}-V(s^{(t)}))+H(\pi(s^{(t)}))
    \end{equation}
\begin{equation}
\small
L_{\theta_{v}}=(R^{(t)}-V(s^{(t)}))^{2}
\end{equation}
where the optimization goal for policy network is to maximize the advantage function computed as  $R^{(t)}-V(s^{(t)})$ and the entropy $H(\pi(s_{t}; \theta))$ of policy output. The entropy in the optimization objective aims to increase the diversity of actions, which can encourage the agent to learn more ﬂexible policies.

\section{Experiment}

\subsection{Experiment Settings}

\textbf{Training Dataset}. 
We train our model on the Hollywood2 \cite{marszalek2009actions} movie dataset, which is composed of 12 classes of human actions and 10 classes of scenes. The dataset provides comprehensive realistic movie scenes with challenging settings in video format. 
We generate key frames for 44 videos chosen from the ﬁrst 100 human action videos and resize them to 540 x 900. Each video consists of 40 randomly sampled sets of images with different numbers for composing the ﬁnal collage. A total of \textasciitilde 2,000 image sets are included in the training set, from which our agent learns robust policies from videos with diverse quality.

\textbf{Evaluation Dataset}. 
To prove the scalabilites of the proposed model, we test it on 10 videos from Hollywood2 dataset and three  movies from the MPII Movie Description dataset\cite{lsmdc}(LSMDC3), including "the queen", 'up in the air', 'pride and prejudice'. The LSMDC dataset provides image collections extracted from sequential time clips. To make meaningful collages and meet the needs of real world scenarios, the evaluation image series in LSMDC3 are chosen from the same context scene and duplicated frames are removed within the same time period.  A total of \textasciitilde 400 and 600 image sets are included in the above two test sets, respectively.

\textbf{Implementation Details}. The aesthetic network and the RL network are implemented with PyTorch \cite{paszke2019pytorch} on Ubuntu 16.04. 
  Our aesthetic evaluation is outputted by View Proposal Network pretrained on the CPC dataset \cite{wei2018good}, which is an aesthetic ranker with advanced image aesthetic evaluation accuracy. To stabilize the training process,  the evaluation and deep aesthetic network share the same feature-extracting unit.

During training, the max epoch is set to 50. For stability reasons, a signal function for reward is used the first 20 epochs and removed for the remaining 30 epochs. The max step is set to 12.
We set 32 for batch size and use the Adam \cite{kingma2014adam} optimizer. The learning rate and weight decay are set to $1e–3$ and $1e–5$. In the A2C algorithm, we set the discount factor to 0.99 and the entropy weight to 0.01. We construct the LSTM unit with four layers in the agent and set the weight of the aesthetic score and blank loss to 1 and 0.01 respectively. 

\textbf{Evaluation Metrics.}
  To assess the quality of the collage results comprehensively, the aesthetic score is developed to describe the overall quality of the collage results, which computes as  
\begin{equation}
\small
  F(C)=\sum_{i}^{N} s\left(P_{i}\right) \cdot f\left(P_{i}\right) \cdot \delta\left(\frac{s\left(P_{i}\right)}{s(C)}>\eta\right)
    \label{fml:fc}
\end{equation}
where $s(P_{i})$ and $f(P_{i})$  represent the area and score of the $i$-th patch, respectively; and N denotes the proposal box number satisfying the selection criteria and collage results with more aesthetic box number are indicators of higher aesthetic quality. The $\eta$ is set to 60\% by default in practice.

The evaluation metrics consider the size and quality of different local regions, and the intuition behind the metric here is to assess the global quality of a holistic collage with accumulated local composition quality of sub-collage-parts. With equation \ref{fml:fc}, the high-quality collage accumulates more aesthetic score over local parts, by means of more balanced compositions, less occlusion along boundaries or fewer blending artifacts.


\begin{figure}[t]
  \includegraphics[width=\linewidth]{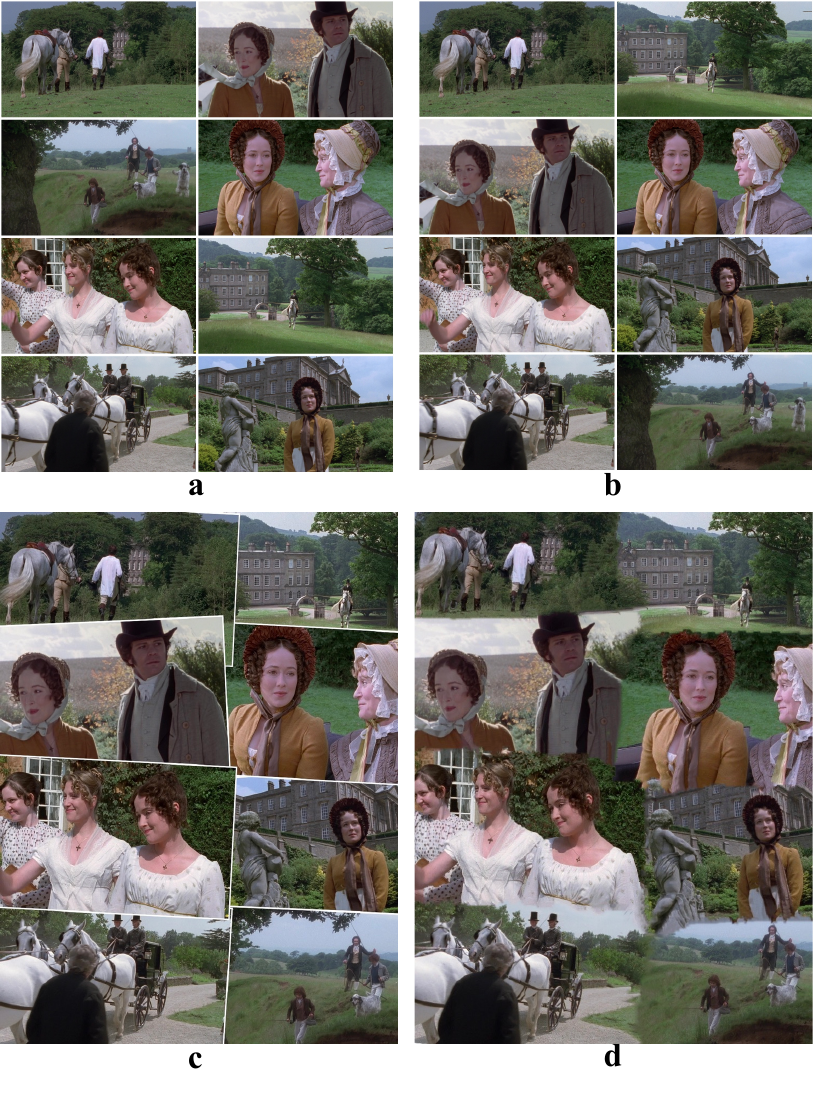}
\caption{Visual comparisons of the impact of the proposed actions on the global layout and local details of the generated collage.
(a) The baseline method is the quick initialization result from Picture Collage which arranges the layout mainly considering salience energy.
(b) The C1 actions transform the global layout and get the aesthetic layout augmented with rule of center.
(c) The C2 actions adjust the local details of individual images and assist to highlight the relevant aesthetic frames, after which the agent generates the aspect ratio specified result (eg. "3:4") as the output.
(d) The blending style could be optionally added to the collage result along the boundaries for the purpose of seamless transition between adjacent images.
}\label{fig:four}
\end{figure}

\begin{table*}[bp]
\begin{center}
\begin{tabular}{|l|ccc|ccc|}
\hline
\multirow{2}{*}{\diagbox[dir=SE, height=2em]{Method}{Input No.}{Evaluation}} & \multicolumn{3}{c|}{Proposal Number} & \multicolumn{3}{c|}{Aesthetic Score} \\ \cline{2-7} 
                  & 6        & 8       & \textless{}15   & 6        & 8       & \textless{}15   \\ \hline
Baseline          & 10.9     & 10.16   & 11.34           & 87.18    & 81.25   & 85.5            \\
\textbf{Ours (w/ C1)}         & 12.31    & 11.49   & 12.53           & 97.51    & 93.81   & 93.72           \\
\textbf{Ours (w/ C2)}      & 13.08    & 12.06   & 13.15           & 101.71   & 95.02   & 98.36           \\ \hline
\end{tabular}
\caption{Aesthetic evaluation on the global layout and local details compared with the baseline method on the Hollywood2 dataset. The second row implies the number of input images, with fixed or unfixed number at one time. The C1 and C2 denote the two categories of proposed action space in Table \ref{tab:action}. }\label{tab:res2}
\end{center}
\end{table*}


\subsection{Quantitative Evaluation}

To assess the effectiveness of the proposed model, we evaluate different methods quantitatively. We examine the effect of our action space design, which adjusts the global layout and local details of a collage in the ﬁrst section, and compare the proposed model’s aesthetic score with that of several competing methods in the second section.

\begin{table}[h]
\caption{Quantitative evaluation for the aesthetic quality of photo collage with different methods generated on the Hollywood2 dataset.}\label{tab:h2}
\centering
\begin{tabular}{|l|c|}
\hline
{Methods}                      & Aesthetic Score      \\ \hline
{Instagram Layout \cite{ins2015instagramlayout}}             & 79.83                \\
{Shape Collage \cite{shp2013shapecollage}}                & 88.22                \\ \hline
{Circle Packing Collage \cite{yu2013content}}               & 63.1                 \\
{Picture Collage \cite{liu2009picture}}              & 83.2                 \\
{AutoCollage \cite{rother2006autocollage}}                  & 103.6                \\ \hline
{\textbf{Ours (w/ AutoCrop)}}  & \textbf{104.8}       \\
{\textbf{Ours (w/o attention)}} & \textbf{106.1}       \\
{\textbf{Ours}}                & \textbf{110.6}       \\ \hline                             
\end{tabular}
\end{table}


\subsubsection{Evaluation of Aesthetic Quality} \

We conduct quantitative comparisons with other existing methods to evaluate the eﬀectiveness of the proposed network.

The competing methods include (a) AutoCollage \cite{rother2006autocollage}, which creates a collage of representative elements from a image set and develops a sequence of optimization steps for collage generation; (b) Circle Packing Collage  \cite{yu2013content}, which partitions a canvas using the importance of regions of interest from input images; (c) Picture Collage  \cite{liu2009picture}, which addresses the photo collage issue with handcrafted energy terms and generates collages through quick initialization and Markov Chain Monte Carlo optimization; (d) Instagram Layout\cite{ins2015instagramlayout}, which is an app developed by Instagram that combines multiple photos into one single image with predefined templates; and (e) Shape Collage\cite{shp2013shapecollage}, which is an automatic photo collage maker that allows to make shape or blending collages of family photos in a harmonious way with more flexible templates. 

To generate results from the same test set in volume, we run a simulation click program on a compiled software to automatically generate the results of AutoCollage, Circle Packing Collage, Instagram Layout and Shape Collage. Given that AutoCollage Touch 2009 has an input number limit, it can only generate collages with more than six input images. As Picture Collage also achieves competitive results, we reimplement its quick initialization process to make comparisons. 

From Table \ref{tab:h2} and  \ref{tab:lsmdc3}, we can see our method achieves consistently higher aesthetic score results than competing methods on both video and image datasets.

\begin{table}[h]
    \caption{Quantitative evaluation of different collage methods on the LSMDC3 dataset.}
    \label{tab:lsmdc3}
      \centering
      \begin{tabular}{|l|c|}
      \hline
      {Methods}                      & Aesthetic Score      \\ \hline
      {Circle Packing Collage \cite{yu2013content}}               & 108.84                 \\
      {Shape Collage \cite{shp2013shapecollage}}                & 114.68                \\ 
      {AutoCollage \cite{rother2006autocollage}}                  & 132.65                \\ \hline
      {\textbf{Ours (w/ AutoCrop)}}  & \textbf{135.11}       \\
      {\textbf{Ours}}                & \textbf{138.86}       \\ \hline                             
      \end{tabular}
\end{table}


\begin{figure*}[bp]
  \includegraphics[width=\textwidth,height=8.2cm]{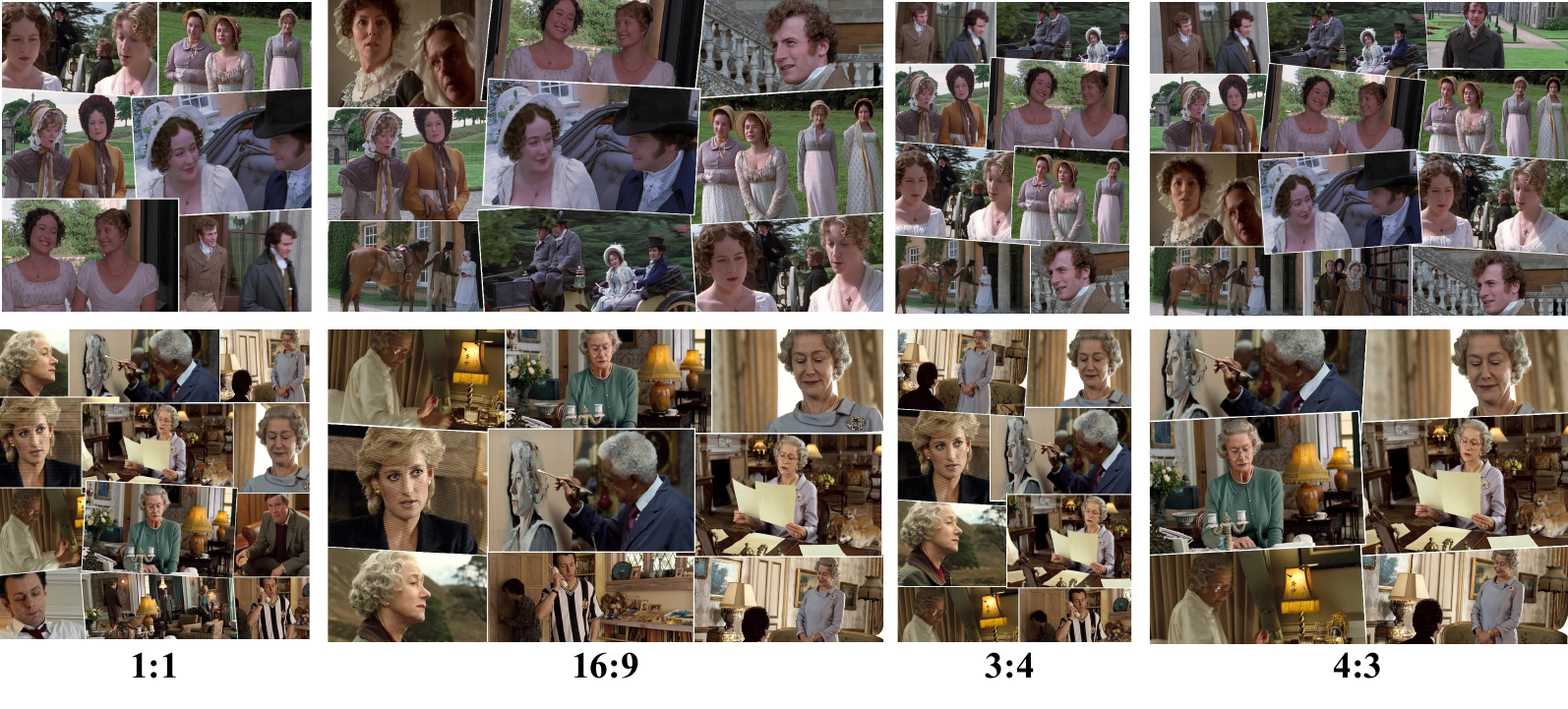}
  \caption{Results of the proposed model on four mainstream aspect ratio specified canvases(i.e., "1:1", "16:9", "4:3", "3:4").  }\label{fig:autocrop}
\end{figure*}

\subsubsection{Evaluation of Action Space Design}\

To evaluate the effectiveness of proposed action space design in Table \ref{tab:action}, we examine the effect of  layout adjustment (C1) and local detail optimization (C2) on the improvement of collage aesthetic quality.  As Picture Collage also generates collages with spatial coordinates, rotation angles, layer indices and performs optimization through handcrafted energy terms, we reimplement quick initialization of picture collage as baseline method and quantitatively compare the quality of the intermediate layout and ﬁnal results of both methods and evaluate them based on their proposal number and aesthetic score. The visual comparison of different action sets are illustrated in Figure \ref{fig:four}.


For global layout adjustment, we use the layout initialization from picture collage as initialization to our network.  We compare the layout results of our proposed network with those of baseline method for multiple different numbered inputs. As shown in Figure \ref{tab:h2} , the agent learns to improve aesthetic quality through image-pair-switch operations and helps create the global layout with increased proposal views, thereby increasing the aesthetic score. 

For local details refinement, both methods are initialized with the same layout for fairness. 
Compared with the baseline method which pays more attention to salience constraints, our method receives feedback from subjective and objective factors. Table \ref{tab:res2} shows that our results can achieve better improvement compared with salience-based optimizations.



\begin{figure*}[tp]
  \includegraphics[width=\textwidth,height=18.2cm]{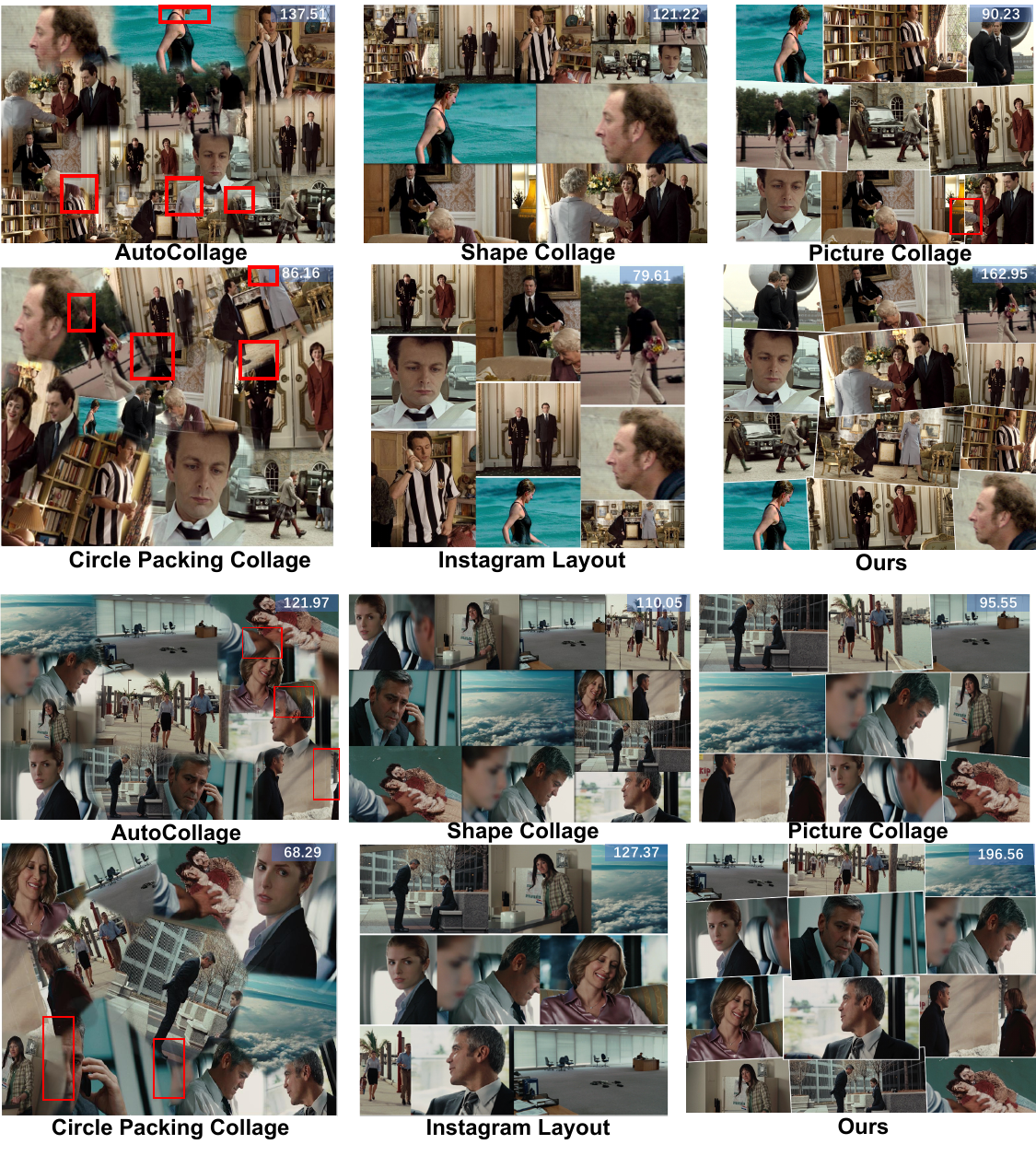}
  \caption{Visual comparisons on the LSMDC3 dataset with competing methods. The aesthetic score using metric in Section 4.1 is labeled in the upright corner of every individual collage from different method.  Artifacts along the boundaries of images are annotated with red boxes.
  In both scenarios, the AutoCollage presents multiple artifacts along the blending borderline, the Circle Packing Collage loses most salient information, and the Picture Collage loses salient information on the canvas border and struggles at highlighting the visual focus. 
  By contrast, our method exhibits the strengths with balanced global layout and preserved local details. Moreover, the salient content is highlighted in the center area while the irrelevant area is decreased with the occlusion.}\label{fig:six}
\end{figure*}

\subsection{Visual Comparisons with Existing Methods}

  
We qualitatively compare our methods against several competitors with three typical collage styles(described in Section 4.2.1), i.e., blending style, overlay style and template style. The visual comparisons are displayed in Figure \ref{fig:six}.

The blending style based methods include AutoCollage \cite{liu2009picture} and Circle Packing Collage \cite{yu2013content}. As shown in Figure \ref{fig:six}, AutoCollage shows multiple artifacts and loses salient information along the boundaries. Also, AutoCollage fails to highlight the visual focus when it comes to complex scenes. Circle Packing Collage loses most salient information and generates confusing boundaries in most cases. Picture Collage \cite{liu2009picture} introduces the overlay style to avoid blending artifacts, however, it is based on salience energy optimization and is unable to generate aesthetic photo collage with good composition quality.
Instagram Layout\cite{ins2015instagramlayout} and Shape Collage \cite{shp2013shapecollage} are template style based methods. The former method generates collage according to the input order and the latter randomly generates photo collage with rich grid templates. However, both methods relies on more user interaction in practice.

Compared to the competing models, our method exhibits the strengths of better composition quality whilst preserving the local details. 
The irrelevant area is greatly decreased with the occlusion while the salience object is well persevered.
Plus, the results of our proposed method have clear boundaries and avoid artifacts.


We visualize the results of the global layout and optimized results of the generated collage in Figure. \ref{fig:four}. With the proper layout (center rules) and detail adjustment, the salient content is highlighted in the center area.
The intuition behind the proposed method is similar to the process of human making collages, thereby making the  automatic generation process interpretable. 
To further improve the scalability, the blending style could be optionally added to the collage result along the boundaries for the purpose of seamless transition between adjacent images.


\subsection{Ablation Study}

To prove the effectiveness of the proposed components, we design ablation experiments to prove the function of the attention fusion module and the AutoCrop module and verify the reasonableness of the evaluation metric.

%
\subsubsection{Attention Fusion Module}\

We examine the influence of attention fusion module on the aesthetic quality of generated collage results and research the effect of the attention mechanism. 

As shown in Table \ref{tab:h2}, after we remove the attention layer, the aesthetic score is observed to decrease by 4.5.  Since the attention layer shifts more focus to the center of a collage, it is an implicit utilization of central rules and thus can enhance the subjective quality of collage results.

    





\subsubsection{AutoCrop Module}\
  
     We investigate the effect of the AutoCrop module on mainstream aspect ratio-speciﬁed canvases to improve the effectiveness in realistic applications.
    
    The AutoCrop module adapts a generated collage with irregular shapes to an aspect ratio-specified canvas. As the AutoCrop module is based on sliding windows powered by the aesthetic network, it can help choose the best view from a raw collage while maximizing the concerned area on the aspect ratio speciﬁed canvas.
  
    To prove the effectiveness, we perform test on four mainstream aspect ratios (i.e., "1:1", "16:9", "4:3", "3:4") and test the capability of the AutoCrop module on both evaluation datasets. The collage results are shown in Figure \ref{fig:autocrop}. For compact aspect ratio, the agent learns to increase the occlusion so as to avoid cropping out salient information, while in the other case the agent learns to decrease the overlay area in order to minimize the blank space.
    
    Results in Table \ref{tab:h2}, \ref{tab:lsmdc3} also show that the concerned content is considerably preserved without being cropped by canvas borders and demonstrate the effectiveness on both video and image datasets.

%
\subsubsection{Evaluation Metric} \

We explore the plausibility of the evaluation metric proposed in Formula \ref{fml:fc}. Specifically, 
 we modify $\eta$  in the fusion process and change it to 50\% and 40\%, respectively.  Although lowering the threshold promises more aesthetic proposals, the increasing number of proposal view brings noisy signals and causes oscillation to the training process and the performance witnesses a reduction by 1.5\% and 4.85\%, respectively. 
Also, increasing the threshold to 70\% reduces the structural information and leads to performance drop by 1.25\%.

\begin{figure}[h]
\centering
\includegraphics[width=0.6\linewidth]{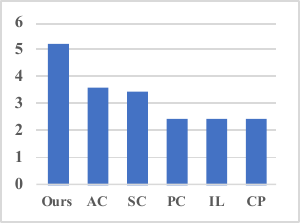}
\caption{User study on different photo collage methods. User study reports results in line with the aesthetic score in Table \ref{tab:h2}, which is a further proof of the effectiveness of proposed evaluation method.} \label{tab:user}
\end{figure}

\subsection{User Study}

Apart from quantitative collage evaluations, to assess the collage methods subjectively, we conduct a user study using a questionnaire. With reference to \cite{rother2006autocollage}, we prepare 24 groups of photo collages by six different methods and invite 20 users not involved in the work to rank the collages with different methods using the same input from top to worst (i.e., 6 to 1) to find visually pleasing collages.
The collages are arranged in random order to avoid biases. Table \ref{tab:user} lists the average scores of each method. The proposed method received high evaluations, which prove its effectiveness. 
Moreover, the user study reports results in line with the aesthetic score in Table \ref{tab:h2}, which is a further proof of the effectiveness of proposed evaluation method.

\section{Conclusion}

In this paper, a novel pipeline for automatic photo collage generation is proposed. Inspired by manual collages, collage generation is decomposed into interpretable steps and modeled as RL for the first time. The attention fusion module embedded in the deep aesthetic network is proposed to overcome the lack of training data and provide a comprehensive feature representation for the photo collage. 
Moreover, the AutoCrop module is proposed to inherently generate an aspect ratio specified collage, which makes the application scenario more flexible.
Experiments on Hollywood2 and LSMDC3 video dataset demonstrate the superiority of the proposed model, and a user study further proves the effectiveness of the subjective evaluation and our method.


%





\ifCLASSOPTIONcaptionsoff
  \newpage
\fi



%

\bibliographystyle{IEEEtran}
\bibliography{reference}

%





\end{document}